\RequirePackage{fix-cm}
\documentclass[smallcondensed]{svjour3}     
\smartqed  
\usepackage{graphicx}
\usepackage{makeidx}  
\usepackage{multirow}
\usepackage{tikz}
\usepackage{pgfplots}
\usepackage{epsfig}
\usepackage{makecell, rotating}
\usepackage{algorithm2e}
\usepackage{comment}
\usepackage{amsmath, amssymb}
\usepackage{bm}

\usepackage{tikz}
\usepackage{pgfplots}
\usepackage{chngpage}

\usepackage{array}
\usepackage{comment}
\usepackage{makecell, rotating}
\usepackage{multirow}
\usepackage{float}
\usepackage{graphicx}
\usepackage{ulem}
\usepackage{filecontents}
\usetikzlibrary{patterns}
\usepackage{setspace}
\usepackage{amsmath}
\usepackage{slashbox}
\usepackage{wrapfig}
\usepackage{graphicx}
\usepackage{placeins}

\usepackage{pgfplotstable}
\pgfplotsset{compat=newest}

\pgfplotstableread{cuhk_model_sel.txt}\cuhkmodelseldata

\pgfplotstableread{traintime.dat}\traintime

\makeatletter
\pgfplotsset{
    /pgfplots/flexible xticklabels from table/.code n args={3}{%
        \pgfplotstableread[#3]{#1}\coordinate@table
        \pgfplotstablegetcolumn{#2}\of{\coordinate@table}\to\pgfplots@xticklabels
        \let\pgfplots@xticklabel=\pgfplots@user@ticklabel@list@x
    }
}
\makeatother

\begin{document}

\title{Learning to Sketch Human Facial Portraits using Personal Styles by Case-Based Reasoning
}
\subtitle{}

\author{Bingwen Jin, Songhua Xu, \and
		Weidong Geng
}

\institute{Bingwen Jin \at
              Department of Computer Science and Technology, \\
              Zhejiang University, Hangzhou, China    \\
              \email{jinbw@zju.edu.cn}             \\
              Weidong Geng \at
              Department of Computer Science and Technology,\\
              Zhejiang University, Hangzhou, China  \\
              \email{gengwd@zju.edu.cn} \\
              Songhua Xu \at
              Department of Information Systems \\
              New Jersey Institute of Technology \\
              \email{songhua.xu@njit.edu}
}

\date{Received: date / Accepted: date}

\maketitle

\begin{abstract}
    This paper employs case-based reasoning (CBR) to
    capture the personal styles of individual artists and generate the
    human facial portraits from photos accordingly.
    For each human artist to be mimicked, a series of cases are firstly built-up from her/his exemplars of source facial photo and hand-drawn sketch, and then its stylization for facial photo is transformed as a style-transferring process of iterative refinement by looking-for and applying best-fit cases in a sense of style optimization. Two models, fitness evaluation model and parameter estimation model, are learned for case retrieval and adaptation respectively from these cases. The fitness evaluation model is to decide which case is best-fitted to the sketching of current interest, and the parameter estimation model is to automate case adaptation. The resultant sketch is synthesized progressively with an iterative loop of retrieval and adaptation of candidate cases until the desired aesthetic style is achieved.
    To explore the effectiveness and advantages of the
    novel approach, we experimentally compare the sketch portraits generated by the proposed method with that of a state-of-the-art example-based facial sketch generation algorithm as well as a couple commercial software packages. The comparisons reveal that our CBR based synthesis method for facial portraits is superior both in capturing and reproducing artists' personal illustration styles to the peer methods.
\end{abstract}

\begin{figure*}
    \setlength{\tabcolsep}{0pt}
    \scriptsize
    \begin{tabular}{*{7}{p{0.143\linewidth}<{\centering}}}
    \includegraphics[width=1\linewidth]{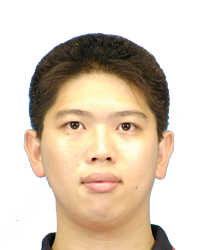}&
    \includegraphics[width=1\linewidth]{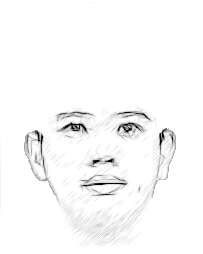}&
    \includegraphics[width=1\linewidth]{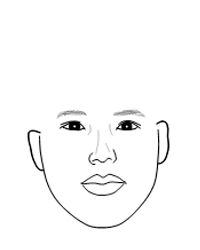}&
    \includegraphics[width=1\linewidth]{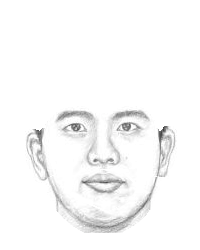}&
    \includegraphics[width=1\linewidth]{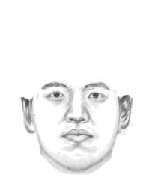} &
    \includegraphics[width=1\linewidth]{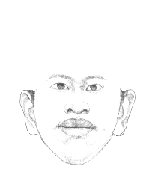} &
    \includegraphics[width=1\linewidth]{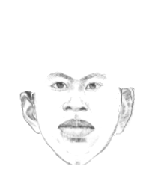} \\
    Input Facial Photo & Peer Method~\cite{akvis2012} & Peer Method~\cite{siyanhui} & Human Artist I & Our Algorithm for Artist I & Human Artist II & Our Algorithm for Artist II
    \end{tabular}
  \caption{Comparison between facial portraits in sketching generated by two commercial software packages (Akvis Sketch~\cite{akvis2012} and a plug-in for Microsoft Outlook 2003~\cite{siyanhui}), two human artists, and our method mimicking the illustration styles of the two artists respectively. In this experiment, our method first learns the personal sketching styles of the two human artists and then generates the sketch portraits for the same input facial photo using the two captured illustration styles respectively.}\label{fig:FrontRslts}
\end{figure*}

\section{Introduction}\label{sec:intr}

Image-based artistic rendering (IB-AR) algorithms often rely on manually-encoded heuristics to emulate specific artistic illustration styles~\cite{Kyprianidis2013}. These heuristics can be classified into three categories~\cite{Kyprianidis2013}: stroke-based rendering for image approximation~\cite{lu2010interactive}, region-based techniques~\cite{zhao2010sisley}, and image processing (filtering)~\cite{papari2007artistic}~\cite{papari2009continuous}. However, it is difficult for artist to explicitly depict the rules of artistic illustration because these rules are often subconsciously exercised and are not always expressible verbally or symbolically. Therefore, example-based artistic rendering (EBAR) is proposed to generate stylized sketch,
which is more popular and more effective since the styles are implicitly embodied in the examples, as it is natural to specify artistic styles by showing a set of examples~\cite{Chen2004}. 

Existing EBAR methods can be broadly classified as model-based and model-free methods. The model-based methods embed prior knowledge about artistic rendering styles into generative models that carry a set of process control parameters. For example, Lu et al.~\cite{Lu2012Pencil} built-up parametric histogram models on the tone distribution inside the sketch example. The resultant sketch was then generated via tone-adjusting by this model according to the tone map in the source image.
Tu et al.~\cite{Tu2010} learned novel direct combined models from the joint distributions of the feature vector pairs of input facial image vs neutral facial shape, neutral vs exaggerated facial shape, and facial sketch vs combination of input facial image and exaggerated facial shape, and then the system optimally synthesized the corresponding output sketch by applying the MMSE criterion~\cite{wu2004automatic} within the combined eigenspace.
The model-free methods synthesize the resultant style-preserving illustrations directly by the correlation relationship between pixels, patches, or strokes etc., in the exemplars. Wang et al.~\cite{Wang2009} divided facial region into overlapping patches from the source photo, and for each patch, located a similar photo patch from examples and collected its corresponding sketch for the synthesis of the resultant sketch smoothly.
Zhao et al.~\cite{Zhao2011} built a dictionary of stroke templates of oil painting portrait with complete information of artists' stroke-by-stroke drawing processes. Their method painterly rendered new portrait by reusing brush strokes from a matched template in the dictionary in terms of facial shape and color in the source image.
Model-based methods can generate diverse new illustrations with parametric styles through generalization on exemplars. However subtle and unique artistic characteristics related to individual artist are somewhat lost while modelling the stylization.
Model-free methods naturally incorporate the visual correlation in examples into the resultant sketch. But the richness of resulting sketches are usually limited by that in the given examples.

In the community of artificial intelligence, case-based reasoning (CBR) solves a new problem instance by first recalling one or multiple similar, previously solved problem instances and then reapplying the known solutions, often with adaptations according to the new context, to address the new problem instance~\cite{aamodt1994case}. From the point of the view of problem solving, the exemplars, a photo and its corresponding artistic illustration, in EBAR are very similar to the cases in CBR. This motivates us to employ case-based reasoning to generate the stylized human facial portraits for an individual artist. Fig.~\ref{fig:FrontRslts} shows two facial sketches automatically created by us using the styles learned from two artists respectively. Sketches generated by two commercial software packages are also presented for comparison.

From a technical perspective, the major challenges for CBR-based stylization of portrait lie in case retrieval and adaptation respectively. Case retrieval in stylization is complex, and goes beyond selecting a best example for a given input as that in~\cite{Wang2009}, because portrait sketching practice of human artists is usually a multi-step and progressive painting process, in which the case retrieval may occur in multiple steps, accounting for not only the given input, but also the current interim sketch and the desired resulting sketch in his/her mind. However, during the phase of looking for the suitable cases, the expected resultant sketch is not available yet for the time being. Case adaptation for stylization is to transfer the style-related correlations in the selected cases into the current sketching. Although existing style-transferring approaches~\cite{Tu2010}~\cite{Wang2009}~\cite{Zhao2011} can be utilized, its key-point is how to automate the case adaptation for stylization, especially when multiple cases are applicable, since automatic case adaptation is necessary to enable CBR systems to function autonomously and to serve naive as well as expert users~\cite{leake1995learning}.

We address these challenges with an approach specific to personalized stylization for sketching faces. First, we designed an iterative pipeline as an overall algorithmic mechanism for CBR-based stylization, in which the current sketch is refined iteratively by the best-fit case until the desired sketch appears. Second, motivated by human artist¡¯s practice of forming a desired sketch in mind before drawing, we train-up a predictive model from cases for each artist to hypothetically ¡±create¡± them. The model evaluates the fitness of each candidate case in terms of the visual similarity to the presumed "resultant sketch". The best-fit one is selected to sketch current face.
Furthermore, in order to automate the case adaptation, a parameter estimation model is learned for each artist in advance, which will automatically assign the appropriate values to parameters for case adaptation, while getting a new case for refinement of resulting sketch.

To summarize, our paper makes the following contributions:
\begin{itemize}
  \item A novel case-based reasoning algorithmic pipeline to iteratively stylize the human portraits by exemplars, producing rich portraits and preserving personal styles well. 

  \item A predictive framework integrated with generate-and-evaluate mechanism is newly proposed for best-fit case retrieval with capability of hypothetically ¡±creating¡± the desired sketch in the mind of artist, which greatly facilitate the evaluation of best-fit cases.

  \item  Innovatively learning parameter estimation model for each artist which enables an automatic case adaptation for stylized sketching of facial photo. 

\end{itemize}

\section{Related Work}

A multitude of image-based artistic rendering (IB-AR) techniques have been proposed. Kyprianidis et al.~\cite{Kyprianidis2013} gave an in depth survey of IB-AR techniques. Here we merely focus on example-based artistic rendering (EBAR), which can be roughly categorized into two classes: model-based and model-free methods.

Model-based methods acquire prior knowledge of artistic rendering styles from examples and accordingly built-up the stylization models. Besides the model-based approaches in~\cite{Lu2012Pencil}~\cite{Tu2010} (see Section~\ref{sec:intr}), Reinhard et al.~\cite{reinhard2001color} modeled color style of a source image through the means and standard deviations along each of the three axes in l$\alpha \beta$ color space, and then imposed the means and standard deviations onto the target image, transferring the color style to target image.
Liang et al.~\cite{Liang2002} models facial pexaggeration style through analyzing the correlation between the image caricature pairs using partial least-squares (PLS).
The model-based hatching in~\cite{Kalogerakis2012} trained a mapping from the features of input 3D object to hatching properties. A new illustration was generated in target style according to predicted properties. The aforementioned model-based methods can generate diverse new illustrations with parametric styles through generalization on exemplars. However subtle and unique artistic characteristics related to individual artist are somewhat lost while modelling the stylization.

Model-free methods generate new artistic illustrations directly by reusing the correlation relationship provided by all exemplars.
Hertzmann et al.~\cite{Hertzmann2001} proposed image analogies algorithm which reuse a source image A and artistic depiction of that image A' to synthesize an artistic illustration B' for a new image B in pixel level. Each new pixel is synthesized by reusing the pixel in exemplar that best matches the pixel being synthesized. Later, an extension of this algorithm incorporated image gradient direction to better preserve object shape of the target image~\cite{Lee2010DTT}.
Example-based stippling~\cite{Kim2009} proposed a texture similarity metric based on gray-level co-occurrence matrix, aiming at generating stipple textures that are perceptually close to input samples. Their reuse of examples merely accounted for pixels related to stipple primitives.
Besides pixel-level reusing of examples, patch-level reusing are also proposed, which often divide original image exemplars into patch exemplars~\cite{Wang2009}.
Liu et al.~\cite{Liu2005} took the similar approach in~\cite{Wang2009}, and retrieved multiple exemplars for each photo patch of a new face and synthesize a sketch patch by linearly blending sketch patched in the candidate exemplars.
Wang et al.~\cite{wang2013transductive} improved this approach by defining a probabilistic model to optimize both the reconstruction fidelity of the input photo and the synthesis fidelity of the target output sketch. Song et al.~\cite{song2014real} further extended it through a Spatial Sketch Denoising method.
Moreover, the face image can also be decomposed into patches in terms of anatomical structure of human face as that in~\cite{Chen2004}~\cite{Gao2009}~\cite{Min2007}~\cite{zhang2014data}.
However, the style of a rendition is mainly embodied in the stokes of a sketch, instead of pixels or patches in an image. Therefore stroke-based reusing in EBAR are also investigated~\cite{Zhao2011}.
Berger et al.~\cite{berger2013style} presented an approach which reuse real artists' strokes to the image. To retain specific styles, they composed a stroke library for each artist, and "cloned" the relevant strokes that matched to the detected edges in the source image.

\section{Method Overview}

\begin{figure}
  \center
  \includegraphics[width=0.5\linewidth]{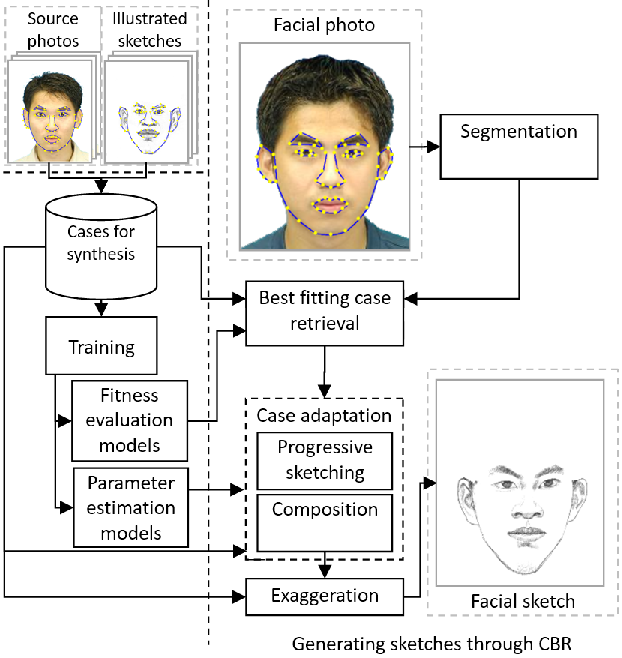}
  \caption{System overview. Feature points on images are represented by yellow dots.}
  \label{fig:overview}
  \includegraphics[width=0.3\linewidth]{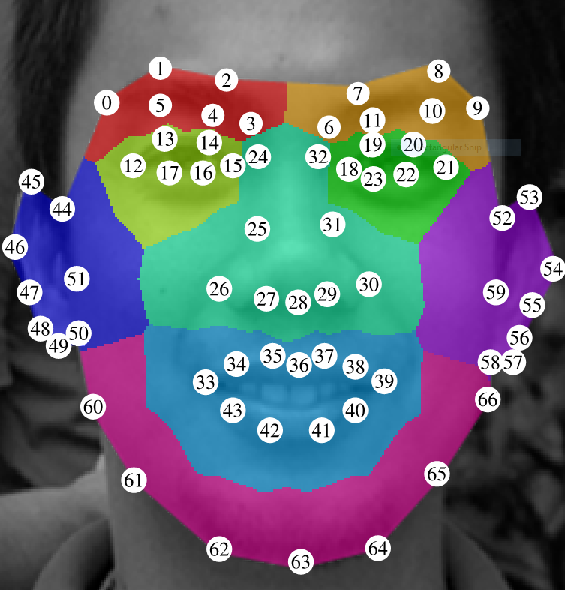}
  \caption{Facial feature points and their corresponding color-coded facial regions.}\label{fig:FFP}
\end{figure}

Fig.~\ref{fig:overview} gives an overview of our CBR-based sketch synthesis framework that produces visually superior results than existing synthesis methods. Given exemplars of source facial photo and stylized sketch hand-drawn by a human artist, its preprocessing phase for case-based reasoning proceeds in three steps: 1) we generate sketch Synthesis (STS) cases and construct a library for them; 2) Fitness Evaluation (FE) models for the artist are trained up from the STS cases; 3) Parameter Estimation (PE) models are learned for automatic case adaptation.
During the runtime phase, we use STS cases, FE models, and PE models to iteratively synthesize a new facial sketch via CBR for each newly given facial photo.

The best-fit case retrieval is carried out by a predictive framework of generate-and-evaluate mechanism. From the point of view of style imitation, best-fit case should be the one that maximally preserves the human artist's style in resulting sketch. However, the resulting sketch is to-be-generated, and  sketch is not at-hand yet, which arises the problem of unknown "ground truth". Our solution is to hypothesize there is always a ground truth in the mind of a human artist, while drawing the sketch. In fact a human artist can easily select the best-fit case by the desired sketch in his/her mind. Therefore we propose a predictive framework embedded with generate-and-evaluate mechanism, in which each candidate case is implicitly applied on the sketch of current interest. FE model is trained to rank its fitness in terms of the similarity between the optimally generated sketch and the one mentally imagined in human artist's mind.

Our case adaptation for sketch synthesis is implemented through a blending operator, whose parameter setting should maximize the similarity between generated sketch and its hypothesized groundtruth. An explicit parameter searching for optimization is time-consuming, therefore we learn a PE model to automatically configure the blending parameter, i.e. to generate a new sketch, we blend the stylized sketch from the best-fit case and the sketch of current interest with the parameter assigned by PE model automatically.

During the phase of iteratively synthesis, we firstly segment the entire facial region into multiple regions according to the anatomical structure of a human face (see Fig.~\ref{fig:FFP} for facial regions painted with different colors). Then for each segmented region,
the resultant sketch is synthesized progressively with an iterative loop of retrieval and adaptation of candidate cases until the desired aesthetic style is achieved.
This is motivated by the human artist's multi-step, progressive portrait sketching practice. FE model guides case retrieval, while PE model automates case adaptation. At last, the overall sketch for the entire human face is composed and formed globally from sketches of these local facial regions according to the relative spatial layout based on facial feature points.

Moreover, a significant characteristics of facial sketch is exaggeration. With our CBR's framework, simulation of the exaggeration is easily embedded as a post-processing step. After our iterative synthesis process, the exaggeration field is generated from STS cases, and then applied on the synthesized sketch to imitate exaggeration style of a human artist.

\section{Building up Cases}\label{sec:datapreparation}

\begin{figure}
  \center
  \includegraphics[width=0.8\linewidth]{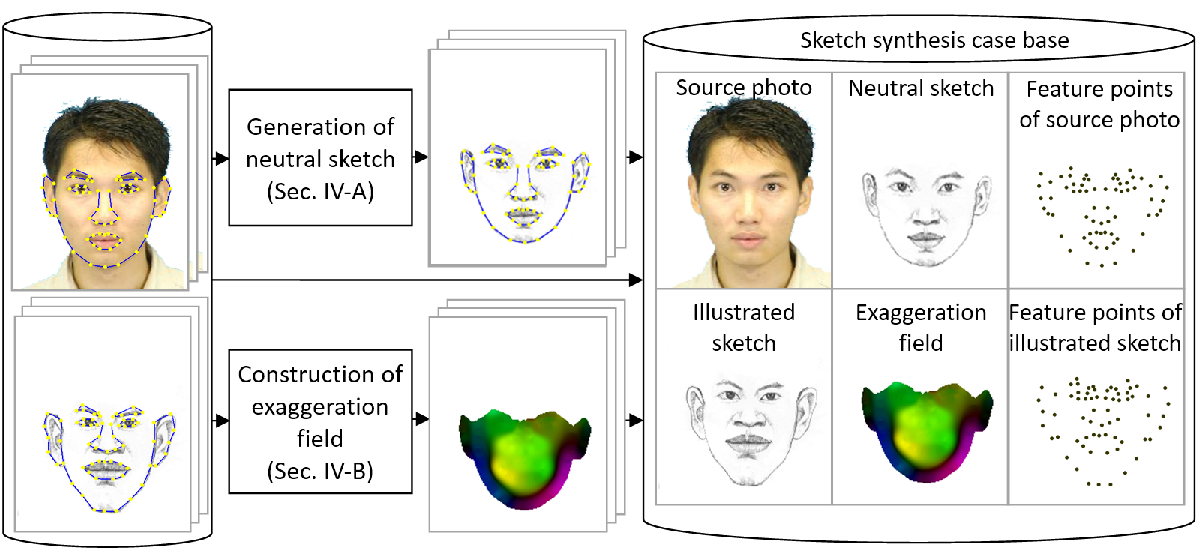}
  \caption{Pipeline of building up cases. Feature points on images are represented by yellow dots.}\label{fig:offline_workflow}
\end{figure}

To build up STS cases (see Fig.~\ref{fig:offline_workflow}), we start from a set of paired source photo and its corresponding stylized sketch, $\Pi_l=\{(I_1, S_1),(I_2, S_2),\cdots,(I_i, S_i),\cdots\}$, where $(I_i, S_i)$ is the $i$-th sample composed of a front-view facial photo $I_i$ and its corresponding stylized sketch $S_i$. All sketches in $\Pi_l$ are illustrated using the same style $l$ as denoted by its subscript. The cardinality of $\Pi_l$, $|\Pi_l|$, is the number of sample pairs inside it.

A STS case consists of 6 components: source photo $I_i$, stylized sketch $S_i$, neutral sketch $S_i'$, exaggeration field $V_i$, and two feature point sets, $Q(I_i)$ and $Q(S_i)$, of $I_i$ and $S_i$ respectively. Each source photo merely has one human face in front view. Stylized sketch is hand-drawn by human artist. Neutral sketch is the one neutralized by removing the exaggeration from the stylized sketch. Neutral sketch will be identical to stylized  sketch if there is no exaggeration in it. Exaggeration field is a 2D vector field representing the displacement of relevant pixels due to exaggeration. Each feature point set has 67 pre-defined points (see Fig.~\ref{fig:FFP}). One is from source photo, and the other one is from stylized sketch.
Formally, we denote a STS case $r_i$ as:
\begin{equation}
  r_i = (I_i, S_i, S_i', V_i, Q(I_i), Q(S_i)).
\end{equation}

A well-established practice of representing the geometry of a facial image is based on the spatial layout of facial feature points. By this principle, the geometry of $I_i$ and $S_i$ are respectively represented by $Q(I_i)$ and $Q(S_i)$. $Q(I_i)$ is a set of 67 pre-defined facial feature points extracted from $I_i$ (see Fig.~\ref{fig:FFP}) by the active shape model (ASM)~\cite{Milborrow2008}. Each feature point is uniquely assigned to one of the nine aforementioned facial regions. Similarly, $Q(S_i)$ is a set of feature points extracted from $S_i$, which are also detected by ASM~\cite{Milborrow2008}. The index numbers of feature points in the two sets $Q(I_i)$ and $Q(S_i)$ as the same. That is, points with the same index number presumably depict the same location on a face, assuming a perfect positional alignment between $I_i$ and $S_i$. However, facial elements sketched by artists sometimes deform geometrically and/or positionally from their counterparts in the corresponding facial photo, partly due to exaggeration. To get the correct matching between them, manual adjustment is often required.

More details about exaggeration field, features, and neutral sketch are given below.

\subsection{Exaggeration Field}\label{sec:CCSE}

Exaggeration field (EF) is a matrix of two-dimensional vectors. The dimension of the matrix is the same as the size of the EF's corresponding image. Each pixel in the image has a counterpart two-dimensional vector that represents the pixel's horizontal and vertical displacements from its position in the neutral sketch to the corresponding position in the deformed sketch with exaggeration.

The EF representing the exaggeration in $S_i$ is denoted as $V_i$. To generate $V_i$, the positions of feature points in $Q(S_i)$ and $Q(I_i)$ are aligned, and the geometric image transformation from $I_i$ to $S_i$ is computed by the image deformation algorithm based on Moving Least Squares (MLS)~\cite{Schaefer2006}. The user can also manually modulate the transformation if needed.
In MLS algorithm, a denser set of displacement vectors $V_i$ are interpolated by the translations between feature points $Q(I_i)$ and $Q(S_i)$. In our prototype implementation, 50000 two-dimensional vectors will be derived in an EF for an image of 200 by 250 pixels.

\subsection{Features}

Given a facial image (sketch or photo) and its feature points, each pixel in the image inside the outline of facial feature points and the facial image is segmented into nine regions by its nearest pre-defined facial feature point by the $L_2$ distance.
Let $\phi_i$ be one of the nine facial regions in $I_i$; $\psi_i$ be $\phi_i$'s corresponding area in $S'_i$; $region(\cdot)$ be a region identification function whose input and output are a facial region and an index representing region type respectively.
Without loss of generation, let $\phi_i$ be the $1$st facial region in $I_i$, which means $region(\phi_i) = 1$.
We denote $Q(\phi_i)$ and $Q(\psi_i)$ as the feature point sets of $\phi_i$ and $\psi_i$ respectively.

Let $\pmb{\tau}_{\textrm{photo}}(\phi_i, Q(\phi_i))$ be a vector of features for $\phi_i$, which includes four types of features, i.e. $\pmb{\tau}_{\textrm{photo}}(\phi_i, Q(\phi_i)) = \big(\pmb{\tau}_{\textrm{surf}}(\phi_i), \pmb{\tau}_{\textrm{gray}}(\phi_i), \pmb{\tau}_{\textrm{dir}}(\phi_i), \pmb{\tau}_{\textrm{context}}(Q(\phi_i))\big)$. Similarly, we extract a set of features for characterizing $\psi_i$, i.e. $\pmb{\tau}_{\textrm{sketch}}(\psi_i, Q(\psi_i))$.

The features used to characterize photo region $\phi_i$ are:
\begin{itemize}
  \item The feature vector $\pmb{\tau}_{\textrm{surf}}(\phi_i)$ is a SURF descriptor~\cite{Bay2006_surf}. For a photo patch $\phi_i$, we extract a $64$ dimensional SURF feature vector to describe the distribution of Haar-wavelet responses in the photo patch.
  \item The feature vector $\pmb{\tau}_{\textrm{gray}}(\phi_i)$ is a normalized histogram on the gray value of pixels in $\phi_i$. We empirically set the histogram dimension to 32 in our implementation.
  \item The feature vector $\pmb{\tau}_{\textrm{dir}}(\phi_i)$ is a normalized histogram on the gradient directions of pixels in $\phi_i$ where the gradient direction of each pixel is calculated using the Sobel operator~\cite{MachineVision}. We also empirically set the histogram dimension to 32.
  \item The feature vector $\pmb{\tau}_{\textrm{context}}(Q(\phi_i))$ is composed of simplified shape context descriptors~\cite{Belongie2002}, which is a log-polar histogram of the coordinates of the remaining points measured using a reference point as the origin. Before generating shape context descriptors, we sample the outline of $Q(\phi_i)$ with uniform spacing, resulting in 100 vertices. Then, for each feature point in $Q(\phi_i)$, a shape context descriptor is computed with 1 and 8 bins for $\log{r}$ and $\theta$ respectively. Therefore, the dimensionality of $\pmb{\tau}_{\textrm{context}}(\phi_i)$ is $8|Q(\phi_i)|$.

\end{itemize}

These features depict the high-level visual characteristics because exaggerated sketch illustrations usually focus more on the salient visual features of an object or scene~\cite{Mish1994} rather than the details.

\subsection{Neutral Sketch}\label{sec:IUS}
\begin{figure}
    \center
    \scriptsize
    \begin{tabular}{ccc}
        \includegraphics[width=0.1\linewidth, trim=15 15 15 15, clip=true]{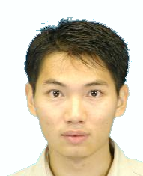} &
        \includegraphics[width=0.1\linewidth, trim=15 15 15 15, clip=true]{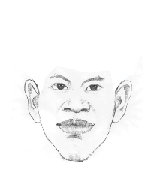} &
        \includegraphics[width=0.1\linewidth, trim=15 15 15 15, clip=true]{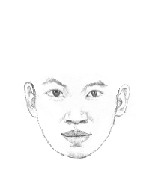} \\
        \includegraphics[width=0.1\linewidth, trim=10 10 20 20, clip=true]{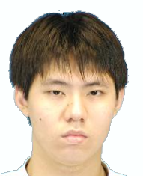} &
        \includegraphics[width=0.1\linewidth, trim=10 10 20 20, clip=true]{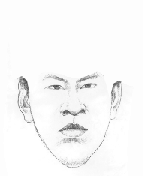} &
        \includegraphics[width=0.1\linewidth, trim=15 10 15 20, clip=true]{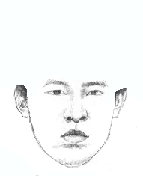} \\
        Photo & \makecell{Artist illustrated \\sketch} & Recovered sketch
    \end{tabular}
    \caption{Two instances of the recovered sketches after removing exaggeration.}
    \label{fig:infer_sketch}
\end{figure}

To generate the neutral sketch, we also use MLS algorithm to derive the geometric image transformation from $S_i$ to $I_i$, which is similar with the method introduced in Sec~\ref{sec:CCSE}. By applying this transformation to $S_i$, we manage to remove the exaggeration from the sample sketch illustration $S_i$, resulting in an neutral sketch $S'_i$. As a byproduct, the method also produces the new positions of the 67 pre-defined facial feature points in $S_i'$, which are denoted by a new feature point set $Q(S_i')$. 
Fig.~\ref{fig:infer_sketch} shows two restored sketches from the stylized facial portraits.

\section{Fitness Evaluation Model for Case Retrieval}

In case retrieval, the fitness is evaluated in terms of the visual similarity between the generated sketch and the groundtruth sketch for the same input facial photo illustrated by the target human artist. The metric for visual similarity assessment is based on the normalized mutual information~\cite{Modat2010FFD}, partly due to mutual information neither depends on any assumption of the data~\cite{thevenaz2000} nor requires the extraction of additional features, such as edges and corners, a process of which may introduce additional geometrical errors~\cite{maes1997}. 

For each facial region, we train one FE model. Let the FE model trained for the $1$st facial region be $f_\textrm{FE}^1$. Its output is fitness quality $\theta_{x,i}$ between a case $r_i$ from our case set and a case $r_x$ which could be another case in our case set or a new case. Its input is a composite feature vector computed from $r_i$ and $r_x$ which is defined as $(\pmb{\tau}_{\textrm{photo}}(\phi_x, Q(\phi_x)), \pmb{\tau}_{\textrm{photo}}(\phi_i, Q(\phi_i)), \pmb{\tau}_{\textrm{sketch}}(\psi_x, Q(\psi_x)))$.
Then we have
\begin{eqnarray}
    \theta_{x,i} &=& f_\textrm{FE}^1(\pmb{\tau}_{\textrm{photo}}(\phi_x, Q(\phi_x)), \pmb{\tau}_{\textrm{photo}}(\phi_i, Q(\phi_i)), \pmb{\tau}_{\textrm{sketch}}(\psi_x, Q(\psi_x))). \label{eq:FE}
\end{eqnarray}

\subsection{Training Data Generation}\label{sec:gendata4fe}

To generate training data for $f_\textrm{FE}^1$, we start from a case set $R = \{r_i|i=1, 2, \cdots, |\Pi_l|\}$, which is randomly divided into 10 equal-sized sub-sets $\{\mathcal{R}_k|k=1,2,\cdots,10\}$. Sketches in one of the sub-sets will be used as the ground truth while the remaining 9 sub-sets are used to generate the sketch results, which will then be cross-validated using the groundtruth sub-set. This process will be performed 10 folds. Therefore, each case will be used once as a groundtruth case. Without loss of generation, let current groundtruth case set be $\mathcal{R}_1$, and remaining case set be $\bar{\mathcal{R}}_1 = \{\mathcal{R}_k|k=2,3,\cdots,10\}$, where $|\bar{\mathcal{R}}_1| = 9|\mathcal{R}_1| = \frac{9}{10}|R|$.

Formally, given a STS case set $\bar{\mathcal{R}}_1$ and a photo region $\phi_x$ from $\mathcal{R}_1$, we identify and adapt one or multiple cases to iteratively synthesize a neutral sketch region to maximally approximate the groundtruth $\hat{\psi}_x$. Let $\psi_x^{\textrm{final}}$ and $\psi_x^{(h)}$ respectively be the final resultant sketch and the interim sketch after the $h$-th iteration. 

In the first iteration, the algorithm selects a case $\bar{r}_x^{(1)}$ whose source photo region $\bar{\phi}_x^{(1)}$ appears visually closest to $\phi_x$ in terms of the normalized mutual information~\cite{Modat2010FFD}. The sketch in the selected case is then denoted as $\psi_x^{(1)}$.

For the $h$-th iteration, the method searches in $\bar{\mathcal{R}}_1$ for a best-fit case $\bar{r}_x^{(h)}$. For each candidate case $r_i$ in $\bar{\mathcal{R}}_1$. The adaptation on $\psi_i$ is carried out by blending $\psi_i$ and $\psi_x^{(h-1)}$. The maximum similarity between resultant sketch patch of blending operation and groundtruth $\hat{\psi}_x$ is
\begin{equation}\label{eq:maxtheta}
    \theta_{x,i}^{(h)} = \max_{\omega \in [0, 1]} \theta\Big(\psi_{i,x,\omega}^{(h - 1)}, \hat{\psi}_x\Big),
\end{equation}
where $\theta_{x,i}^{(h)}$ is also represented as the fitness value of $r_i$, $\psi_{i,x,\omega}^{(h - 1)}$ is the resultant sketch patch of blending $\psi_i$ and $\psi_x^{(h-1)}$ with blending weight $\omega$, and $\theta(\psi_{i,x,\omega}^{(h - 1)}, \hat{\psi}_x)$ is the visual similarity between $\psi_{i,x,\omega}^{(h - 1)}$ and $\hat{\psi}_x$ in terms of normalized mutual information~\cite{Modat2010FFD}.

The image blending operator to synthesize $\psi_{i,x,\omega}^{(h - 1)}$ is originally proposed in~\cite{Lee1996}. It generates an inbetween image, $I_t|_{t \in [0,1]}p$, of two input images, $I_0$ and $I_1$, by $I_t = (1 - t)W_0(t, I_0) + tW_1(t, I_1)$, where $W_0$ and $W_1$ are two non-linear warping functions built from all pairs of corresponding feature points between $I_0$ and $I_1$.
It is obvious that $t$ is identical to our $\omega$.
During the process of blending $I_0$ and $I_1$, their feature point sets, $Q(I_0)$ and $Q(I_1)$, are also blended to compute the feature point set for $I_t$.
Formally, we define the blending operation as follows:
\begin{eqnarray}\label{eq:blend}
    \Big(\psi_{i,x,\omega}^{(h - 1)}, Q(\psi_{i,x,\omega}^{(h - 1)})\Big) &=& \oplus\Big(\psi_x^{(h-1)}, Q\Big(\psi_x^{(h-1)}\Big), \psi_i, Q(\psi_i), \omega\Big),
\end{eqnarray}
where $\psi_{i,x,\omega}^{(h - 1)}$ is the resulting image of blending $\psi_i$ and $\psi_x^{(h - 1)}$ with the blending parameter $\omega \in [0, 1]$; $Q(\psi_{i,x,\omega}^{(h - 1)})$ is the resulting set of facial feature points.

A non-linear optimization method~\cite{Powell1998COBYLA} is employed as a solver for Equation~(\ref{eq:maxtheta}).
After computing $\theta_{x,i}^{(h)}$ for all candidate cases, the case with maximum $\theta_{x,i}^{(h)}$ will be selected as $\bar{r}_x^{(h)}$. We denote the maximum $\theta_{x,i}^{(h)}$ as $\bar{\theta}_x^{(h)}$. Resultant sketch patch corresponding to $\bar{\theta}_x^{(h)}$, $\psi_x^{(h)}$, will be involved in the next iteration. This iterative synthesis procedure terminates when $\frac{\bar{\theta}_x^{(h)} - \bar{\theta}_x^{(h - 1)}}{\bar{\theta}_x^{(h - 1)}}$ is less than a threshold (it is set to $1\%$ in our implementation).

All $\theta_{x,i}^{(h)}$ generated in aforementioned iterative process are collected as sample set $\{\theta_{x,i}^{(h)}|r_i\in \bar{\mathcal{R}}_1, h=1,2,\cdots,n_{\textrm{step}}(\phi_x)\}$, where $n_{step}(\phi_x)$ is the number of iterations, and all combinations of $(\phi_x, Q(\phi_x), \phi_i, Q(\phi_i), \psi_x^{(h-1)}, Q(\psi_x^{(h-1)}))$ are collected as training samples. Hence we generate $|\bar{\mathcal{R}}_1|n_{\textrm{step}}(\phi_x)$ samples for $f_\textrm{FE}^1$ after synthesizing a sketch for $\phi_x$. Supposing that the mean value of $n_{\textrm{step}}(\phi_x)$, $\overline{n_{\textrm{step}}(\phi_x)}$, is $5$, and the size of $R$ is $50$, we can generate $10|\mathcal{R}_1||\bar{\mathcal{R}}_1|\overline{n_{\textrm{step}}(\phi_x)} = 0.9|R|^2p\overline{n_{\textrm{step}}(\phi_x)}=11250$ training samples.

\subsection{Learning Fitness Evaluation Model}\label{sec:train}

Given the training samples for $f_\textrm{FE}^1$, we train the fitness evaluation model via regression. To identify optimal regression model for $f_\textrm{FE}^1$, we employ the ten-fold cross validation (CV) technique during the model selection process~\cite{schaffer1993}.

The definitions of $\pmb{\tau}_{\textrm{photo}}(\cdot)$ and $\pmb{\tau}_{\textrm{sketch}}(\cdot)$ indicate that the input to $f_\textrm{FE}^1$ consists of hundreds of features. The dimensionality of the input features is between 528 and 648 wherein the exact dimensionality depends on the specific type of the facial region. Its feature selection is accomplished using the minimal-redundancy-maximal-relevance (mRMR) criterion~\cite{Peng2005}. However, the mRMR method can only identify the most important features by a user-specified number. To optimally get this number, we employ the best-first search algorithm~\cite{kohavi1997wrappers} to look for it through minimizing the CV error.
The best regression model is selected by the minimized CV error derived through performing feature selection for each candidate regression model.

\begin{figure}
    \center
    \includegraphics[width=0.5\linewidth]{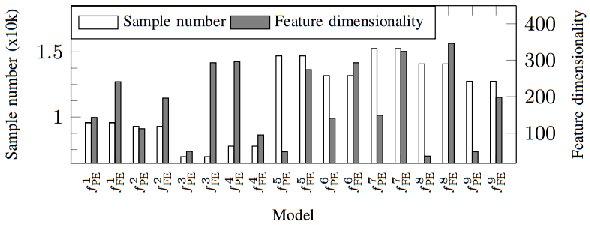}
    \caption{The sizes of the training sample sets for the nine facial regions. The reported feature dimensionality refers to the reduced dimensionality of the selected feature set with the optimally chosen regression models.}\label{plot:DataSets}
    \includegraphics[width=0.5\linewidth]{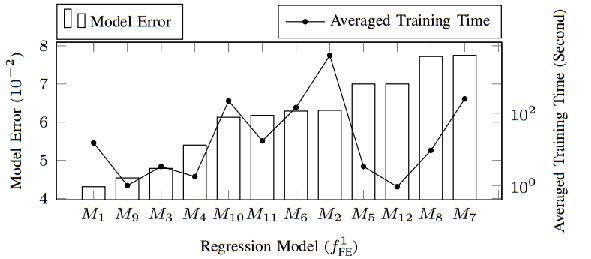}
    \caption{ The cross-validation errors of candidate regression models of $f_\textrm{FE}^1$ for $\Pi_1$. The averaged training time for each model is also shown.}\label{plot:FEModelSel}
    \includegraphics[width=0.5\linewidth]{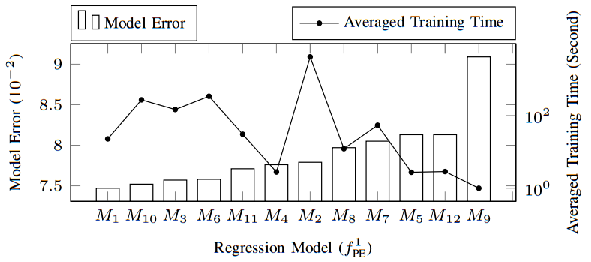}
    \caption{ The cross validation errors of candidate regression models of $f_\textrm{PE}^1$ for $\Pi_1$. The averaged training time for each model is also shown.}\label{plot:ModelSel4PE}
\end{figure}

\begin{table}
\center
\scriptsize
\setlength{\tabcolsep}{3pt}
\label{tab:opt_bp_model}
\caption{Selected optimal regression models for the FE models for datasets $\Pi_1$, $\Pi_2$, and $\Pi_3$ respectively.}
\begin{tabular}{*{10}{|c}|}
    \hline
    \backslashbox{Dataset}{Region} & 1 & 2 & 3 & 4 & 5 & 6 & 7 & 8 & 9 \\
    \hline
    $\Pi_1$ & $M_1$ & $M_9$ & $M_9$ & $M_9$ & $M_9$ & $M_3$ & $M_9$ & $M_1$ & $M_1$  \\
    \hline
    $\Pi_2$ & $M_1$ & $M_1$ & $M_9$ & $M_9$ & $M_1$ & $M_1$ & $M_1$ & $M_1$ & $M_1$ \\
    \hline
    $\Pi_3$ &  $M_1$ & $M_1$ & $M_7$ & $M_7$ & $M_1$ & $M_7$ & $M_1$ & $M_3$ & $M_7$ \\
    \hline
\end{tabular}
\label{tab:opt_fe_model}

\center
\scriptsize
\setlength{\tabcolsep}{3pt}
\caption{Selected optimal regression models for the PE models for datasets $\Pi_1$, $\Pi_2$, and $\Pi_3$ respectively.}
\begin{tabular}{*{10}{|c}|}
    \hline
    \backslashbox{Dataset}{Region}  & 1 & 2 & 3 & 4 & 5 & 6 & 7 & 8 & 9 \\
    \hline
    $\Pi_1$ & $M_1$ & $M_1$ & $M_1$ & $M_1$ & $M_1$ & $M_1$ & $M_6$ & $M_6$ & $M_1$ \\
    \hline
    $\Pi_2$ & $M_3$ & $M_{11}$ & $M_1$ & $M_1$ & $M_3$ & $M_3$ & $M_1$ & $M_9$ & $M_1$ \\
    \hline
    $\Pi_3$ & $M_1$ & $M_1$ & $M_3$ & $M_3$ & $M_1$ & $M_1$ & $M_1$ & $M_1$ & $M_1$  \\
    \hline
\end{tabular}
\label{tab:opt_pe_model}
\end{table}

Fig.~\ref{plot:DataSets} shows the sample number and feature dimensionality in the model selection process based on dataset $\Pi_1$ (see Section~\ref{sec:ExpResults} for details about $\Pi_1$).
Using the Weka toolkit~\cite{Hall2009_weka}, 12 most popular regression models, $\{M_k|k=1,2,\cdots,12\}$, are taken into consideration: bagging regression tree ($M_1$), SVM regression ($M_2$), M5P ($M_3$), regression tree ($M_4$), conjunctive case ($M_5$), M5Cases ($M_6$), isotonic regression ($M_7$), additive regression ($M_8$), KNN ($M_9$), linear regression ($M_{10}$), pace regression ($M_{11}$), and decision stump ($M_{12}$). Let $\epsilon_k$ be the cross-validation errors of the $k$-th candidate regression model $M_k$ for $f_\textrm{FE}^1$, which are applicable for the $1$st facial region. Then $\{\epsilon_k|k=1,2,\cdots,12\}$ are used to select an optimal regression model for  $f_\textrm{FE}^1$. Fig.~\ref{plot:FEModelSel} shows the cross-validation errors of the candidate regression models of $f_\textrm{FE}^1$ for $\Pi_1$, from which we find the optimal one is bagging trees~\cite{Breiman1996}. Table~\ref{tab:opt_fe_model} shows the regression models selected for $\Pi_1$, $\Pi_2$, and $\Pi_3$ (see Section~\ref{sec:ExpResults} for details about $\Pi_2$ and $\Pi_3$), in which we observe that the optimal regression models chosen for FE models may vary across facial regions and datasets.

\section{Parameter Estimation Model for Automatic Case Adaptation}

Case adaptation is carried out by blending the sketch from the retrieved case with the interim sketch from previous iteration. To automate the case adaptation, we train PE models to identify the optimal blending parameter, maximizing the similarity between the resultant sketch of blending and the hypothetic groundtruth sketch.

Let $f_\textrm{PE}^1$ be the PE model trained for the $1$st facial region. In each iteration of sketch synthesis, a new photo region $\phi_x$, a sketch region of current interest $\psi_x$, and a case $r_i$ are given. PE model is to estimate the optimal parameter $\omega_{x,i}$ for blending $\psi_x$ and $\psi_i$. That is:
\begin{eqnarray}
    \omega_{x,i} &=& f_\textrm{PE}^1(\pmb{\tau}_{\textrm{photo}}(\phi_x, Q(\phi_x)), \pmb{\tau}_{\textrm{photo}}(\phi_i, Q(\phi_i)), \pmb{\tau}_{\textrm{sketch}}(\psi_x, Q(\psi_x))). \label{eq:PE}
\end{eqnarray}
The learning of PE model takes the same training samples for FE model. It is worth noticing that an optimal blending weight is also acquired, after solving Equation~(\ref{eq:maxtheta}).
Therefore, the training data for $f_\textrm{PE}^1$ is also accordingly generated as a by-product while preparing training data for $f_\textrm{FE}^1$, which is described in Section~\ref{sec:gendata4fe}.

We use the same learning method introduced in Section~\ref{sec:train} to train PE model. Fig.~\ref{plot:DataSets} also shows the sample number and feature dimensionality in the model selection process for PE models. Fig.~\ref{plot:ModelSel4PE} shows the cross-validation errors of the candidate regression model for $f_\textrm{PE}^1$ of $\Pi_1$. Table~\ref{tab:opt_pe_model} shows the regression models selected for $\Pi_1$, $\Pi_2$, and $\Pi_3$ (see Section~\ref{sec:ExpResults} for details about $\Pi_1$, $\Pi_2$ and $\Pi_3$), in which we can also observe that the optimal regression models chosen for PE models vary across facial regions and datasets.

\section{Synthesizing the Desired Sketch via CBR}

\begin{figure}
    \center
    \scriptsize
    \setlength{\tabcolsep}{3pt}
    \begin{tabular}{cccc}
        Use 1 Case &  & Use Multiple Cases & Groundtruth \\
        \makecell{\includegraphics[width=0.15\linewidth, trim=0 5 0 0, clip=true]{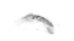}}&
        $\cdots$ &
        \makecell{\includegraphics[width=0.15\linewidth, trim=0 5 0 0, clip=true]{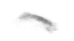}}&
        \makecell{\includegraphics[width=0.15\linewidth, trim=0 5 0 0, clip=true]{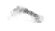}} \\
        1: 0.9024 &
        &
        4: 0.9036 &
        \\
        \makecell{\includegraphics[width=0.1\linewidth, trim=0 15 0 0, clip=true]{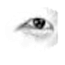}}&
        $\cdots$ &
        \makecell{\includegraphics[width=0.1\linewidth, trim=0 15 0 0, clip=true]{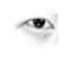}}&
        \makecell{\includegraphics[width=0.1\linewidth, trim=0 15 0 0, clip=true]{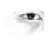}} \\
        1: 0.8163&
        &
        3: 0.8316&
        \\
        \makecell{\includegraphics[width=0.15\linewidth, trim=0 10 0 0, clip=true]{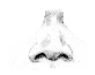}}&
        $\cdots$ &
        \makecell{\includegraphics[width=0.15\linewidth, trim=0 10 0 0, clip=true]{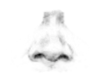}}&
        \makecell{\includegraphics[width=0.15\linewidth, trim=0 10 0 0, clip=true]{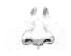}} \\
        1: 0.8472&
        &
        3: 0.8700&
        \\
        \makecell{\includegraphics[width=0.05\linewidth]{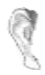}}&
        $\cdots$ &
        \makecell{\includegraphics[width=0.05\linewidth]{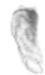}}&
        \makecell{\includegraphics[width=0.05\linewidth]{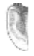}} \\
        1: 0.7329&
        &
        3: 0.7455&
        \\
        \makecell{\includegraphics[width=0.15\linewidth, trim = 0 10 0 0, clip=true]{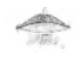}}&
        $\cdots$ &
        \makecell{\includegraphics[width=0.15\linewidth, trim = 0 10 0 0, clip=true]{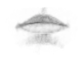}}&
        \makecell{\includegraphics[width=0.15\linewidth, trim = 0 10 0 0, clip=true]{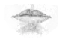}} \\
        1: 0.8865&
        &
        3: 0.9276&
        \\
        \makecell{\includegraphics[width=0.15\linewidth, trim=0 0 0 20, clip=true]{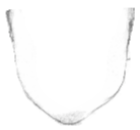}}&
        $\cdots$ &
        \makecell{\includegraphics[width=0.15\linewidth, trim=0 0 0 20, clip=true]{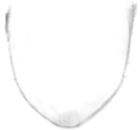}}&
        \makecell{\includegraphics[width=0.15\linewidth, trim=0 0 0 20, clip=true]{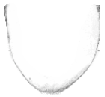}}\\
        1: 0.9378&
        &
        5: 0.9531&
        \\
    \end{tabular}
    \caption{ Comparison between sketch synthesis results using single case (first column) and multiple cases (second column). The third column is the ground truth. The number of the cases used and the similarity with the ground truth image measured by normalized mutual information~\cite{Modat2010FFD} is also given for reference. }\label{fig:cmp_patch_blend}
\end{figure}

\begin{figure}
    \center
    \scriptsize
    \setlength{\tabcolsep}{1pt}
    \begin{tabular}{cccc}
        \scriptsize{Using 1 Case} &  & \scriptsize{Using Multiple Cases} & \scriptsize{Ground Truth} \\
        \makecell{\includegraphics[width=0.15\linewidth, trim=0 20 0 70, clip=true]{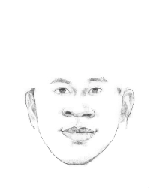}}&
        $\cdots$ &
        \makecell{\includegraphics[width=0.15\linewidth, trim=0 20 0 70, clip=true]{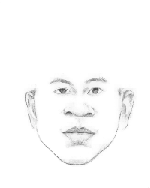}}&
        \makecell{\includegraphics[width=0.15\linewidth, trim=0 20 0 75, clip=true]{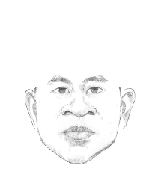}} \\
        \scriptsize{Sim = 0.5697} &
        &
        \scriptsize{Sim = 0.6141} &
        \\
        \makecell{\includegraphics[width=0.15\linewidth, trim=0 20 0 80, clip=true]{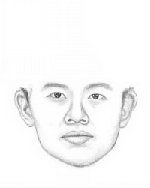}}&
        $\cdots$ &
        \makecell{\includegraphics[width=0.15\linewidth, trim=0 20 0 80, clip=true]{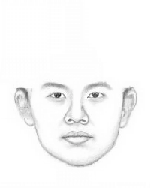}}&
        \makecell{\includegraphics[width=0.15\linewidth, trim=0 10 0 100, clip=true]{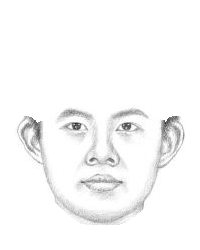}} \\
        \scriptsize{Sim = 0.5175} &
        &
        \scriptsize{Sim = 0.5367} &
        \\
    \end{tabular}
    \caption{ Comparison between the synthesized facial sketches using single case (first column) and multiple cases (second column) for each facial region as compared with the groundtruth facial sketches (third column). "Sim" is its visual similarity with the groundtruth image measured by normalized mutual information~\cite{Modat2010FFD}.}\label{fig:step_wise_gen}
\end{figure}

Given an input facial photo and a set of STS cases of a human artist,
the resultant sketch is synthesized progressively with an iterative loop of retrieval and adaptation of candidate cases until the desired aesthetic style is presented. Fig.~\ref{fig:cmp_patch_blend} and~\ref{fig:step_wise_gen} clearly show that the synthesized sketch using multiple cases is much more close to the ground truth image than the sketch produced using single case in a sense of normalized mutual information~\cite{Modat2010FFD}. Moreover, exaggeration imitation is embedded as a post-processing step.

\subsection{Iterative sketching by Cases}\label{sec:LSGA}

\begin{figure}
    \center
    \includegraphics[width=0.45\linewidth]{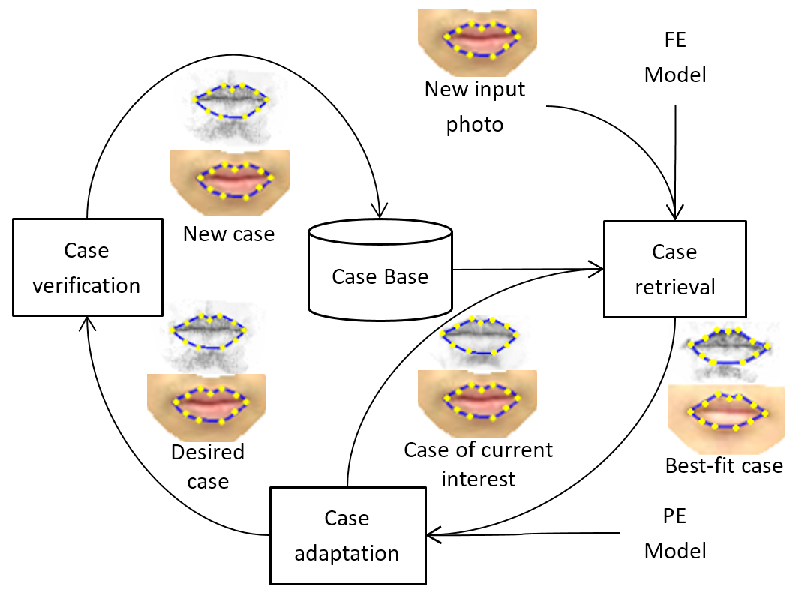}
    \caption{The pipeline of synthesize sketches for individual facial regions iteratively through case-based reasoning. Feature points on images are represented by yellow dots.}
    \label{fig:stroke_synthesis_workflow}
\end{figure}

To synthesize the sketch in a given input facial photo $I_x$, the key point is how to optimally retrieve relevant cases and adaptively fuse them to produce the resultant sketch. Our CBR based synthesis pipeline is illustrated in Fig.~\ref{fig:stroke_synthesis_workflow}, which originates from the general CBR framework~\cite{aamodt1994case}.

Given a new facial region $\phi_x$ from a new photo $I_x$ and a set of STS cases, the process for synthesizing a new sketch $\psi_x$ is very similar to the one in Section~\ref{sec:gendata4fe}. The major difference is that $f_\textrm{FE}^1$ and $f_\textrm{PE}^1$ are already known now. Therefore, $\theta_{x,i}^{(h)}$ and optimal $\omega$ of Equation~\ref{eq:maxtheta} can be directly calculated by $f_\textrm{FE}^1$ and $f_\textrm{PE}^1$ respectively, ignoring the non-linear optimization for  Equation~\ref{eq:maxtheta}.

We are aware of that Liu et al.~\cite{Liu2005} introduced a similar method for finding optimal example sketch patches and their blending parameters during a facial texture synthesis procedure. Their method is based on the local linearity assumption, which searches the most visually similar examples for an input photo region. The optimal blending parameters are identified through minimizing the reconstruction error in terms of the visual similarity between the input photo and the blended result of the selected example photos.
Instead of calculating blending parameters directly from examples, a large amount of training data from a limited number of sketch synthesis cases are generated to train PE models, which allows us to fully utilize available cases for parameter estimation.

The blended image will appear more blurry than the source one, and the post-processing is usually required.  We extend the image analogies algorithm~\cite{Hertzmann2001}, as an example-based image sharpening procedure with a multi-scale autoregression process, which can learn from multiple pairs of example images. More concretely, we use $\{\psi_i|i=1, 2, \cdots, |\Pi_l|\}$ as the ``filtered" examples and apply a Gaussian kernel, whose radius and standard deviation are empirically set to $3$ and $1$ pixels respectively, over each $\psi_i$ to generate its  ``unfiltered" version $\psi_i^{\textrm{smoothed}}$. According to the exemplified mapping relation  $\{(\psi_i^{\textrm{smoothed}}, \psi_i)|i=1, 2, \cdots, |\Pi_l\}$, the sharpened one of $\psi_x^{\textrm{final}}$ is synthesized by image analogy. Fig.~\ref{fig:CmpStrEnhance} shows that its sharpened sketch is much better than the one processed by the high-pass filter-based method~\cite{gonzalez2002digital} in terms of the visual appearance of sketching.

\subsection{Exaggeration}\label{sec:FSE}
\begin{figure}
    \center
    \scriptsize
    \setlength{\tabcolsep}{4pt}
    \begin{tabular}{*{3}{c}}
        \includegraphics[width=0.2\linewidth]{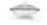} &
        \includegraphics[width=0.2\linewidth]{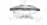} &
        \includegraphics[width=0.2\linewidth]{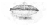} \\
        The blended mouth & By high pass filter & By our Method
    \end{tabular}
    \caption{Comparison with a high-pass filter based method. Our method can present more details than the high-pass filter based method.}\label{fig:CmpStrEnhance}

    \setlength{\tabcolsep}{1pt}
    \center
    \scriptsize
    \begin{tabular}{*{4}{c}}
        \includegraphics[width=0.14\linewidth, trim=0 0 0 40, clip=true]{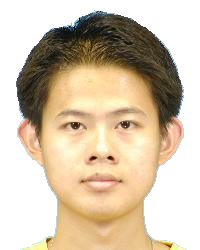} &
        \includegraphics[width=0.14\linewidth, trim=0 0 0 50, clip=true]{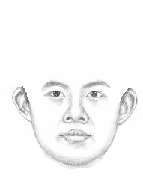} &
        \includegraphics[width=0.14\linewidth, trim=0 0 0 0, clip=true]{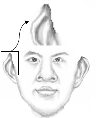} &
        \includegraphics[width=0.14\linewidth, trim=0 0 0 0, clip=true]{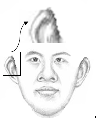} \\
        Photo & (a) & (b) & (c)
    \end{tabular}
    \caption{Deforming a sketch using the exaggeration field and feature point displacement vectors sampled from the exaggeration field. (a) is an initial sketch. (b) is generated by a standard image remapping algorithm~\cite{opencv_library}. (c) is generated by our revised approach. Notice: (b) and (c) are best viewed at 400\% zoom.}
    \label{fig:GenExgSketch}
\end{figure}

Exaggeration is popular in facial sketch illustrations, which is often captured as a 2D transformation from $S_i$ to $S_i'$, represented by exaggeration field $V_i$.

In our CBR framework, a new EF for exaggeration imitation, $V_x$, is generated by a linear system $H$ as $H\big(Q(S_x')\big) = V_x$, where $S_x'$ is a new facial sketch and $Q(S_x')$ is the set of feature points on $S_x'$. 

The linear system has the properties of superposition and homogeneity of degree 1~\cite{Chen1998LST}, therefore $H\big(\sum_{i=1}^{|\Pi_l|}{\omega_i Q(S_i')}\big) = \sum_{i=1}^{|\Pi_l|}{\omega_i H\big(Q(S_i')\big)}$, where $\{Q(S'_i)|i=1, 2, \cdots, |\Pi_l|\}$ is a set of sample inputs of $H$, and $\omega_i$ is the salience weight associated with $S'_i$. Given a new input $Q(S_x')$, we can optimally find a linear combination of sample inputs to $H$ that best reconstructs $Q(S_x')$. That is:
\begin{eqnarray}
    \label{eq:GenEF0}
    \bar\Omega &=& \arg \min_{\Omega} D\Big(Q(S_x'), \sum_{i=1}^{|\Pi_l|}{\omega_i Q(S_i')}\Big), \\
    \label{eq:GenEF1}
     & s.t. & \omega_i \in [0,1], \sum_{i=1}^{|\Pi_l|} \omega_i = 1, \nonumber \\
    H\left(Q(S_x')\right) &\approx& H\Big(\sum_{i=1}^{|\Pi_l|}{\bar{\omega}_iQ(S_i')}\Big) = \sum_{i=1}^{|\Pi_l|}{\bar{\omega}_i V_i},
\end{eqnarray}
where $\Omega = \{\omega_i|i=1, 2, \cdots, |\Pi_l|\}$, $\bar\Omega = \{\bar\omega_i|i=1, 2, \cdots, |\Pi_l|\}$, and $D(\cdot, \cdot)$ is defined as the sum of Squared Euclidean Distances between pairs of corresponding points in two input feature point sets, and $V_i$ is the sample output of $H$ corresponding to $Q(S_i')$ (see Sec.~\ref{sec:CCSE} for the method of generating $V_i$).
According to the index number specified for each feature point (see Fig.~\ref{fig:FFP}), we can naturally establish a pairwise point-to-point correspondence between the two sets of feature points.

We cannot theoretically prove the linear assumption of our method. However, experimentally, the resultant exaggeration by the predicted EF appears highly close to the ones illustrated by the target human artist.

Once $V_x$ is generated, a straightforward approach for exaggerating $S_x'$ is to warp $S_x'$ following the guidance of $V_x$ through a standard image remapping function~\cite{opencv_library}. However, $V_x$ is usually noisy, which would cause unreliable deformations (see Fig.~\ref{fig:GenExgSketch}(b)). Therefore we firstly get rid of the noise from $V_x$ by interpolating the control point displacements~\cite{Schaefer2006}. And then the smoothed $V_x$ is used to warp $S_x'$ through image remapping~\cite{opencv_library}, which generates the exaggerated sketch $S_x$ accordingly (see Fig.~\ref{fig:GenExgSketch}(c)).

\section{Experimental Results}\label{sec:ExpResults}

In our experiments, we prepared 3 sets of facial sketch illustrations, respectively denoted as $\Pi_1, \Pi_2, \Pi_3$ for three human artists respectively. Each set consists of facial sketch portraits illustrated by one human artist, as differentiated through the subscript of $\Pi_i$'s. Every set contains 49 front-view facial photos of 49 different people where each photo is accompanied by its sketch illustrated by the artist. The resolutions of the images are 200 by 250 pixels in $\Pi_1$, 191 by 235 pixels in $\Pi_2$, and 194 by 247 pixels in $\Pi_3$. In particular, for $\Pi_1$, all its facial photos and their sketch illustrations are from the CUHK student dataset~\cite{Wang2009}. For $\Pi_2$ and $\Pi_3$, 39 photos are from $\Pi_1$ and the remaining 10 new human facial photos are newly taken, which are both included in $\Pi_2$ and $\Pi_3$.
For the 49 photos respectively in $\Pi_2$ and $\Pi_3$, we hired two artists to draw 49 facial sketches for each set respectively. The artist for $\Pi_2$ has more than 20 years of professional experiences in creating human facial portraits. The artist for $\Pi_3$ is a PhD student of digital art and design. He has been a freelance illustrator for 7 years and showed his caricature art pieces in a national culture and art expo event. 
In this manual sketch illustration process, we used a 21-inch LCD to display the human facial photo one by one and asked artists to draw sketch illustrations for each of the displayed facial photo on their A4 paper canvas. Our artists were asked to spend as long as they want in creating these sketch illustrations.
After that, all hand-drawn sketches are digitalized by a scanner. The original resolutions of the images in $\Pi_2$ and $\Pi_3$ are 2480 by 3508 pixels. As the image resolution of the released CUHK student dataset is 200 by 250 pixels, we downsample the images in $\Pi_2$ and $\Pi_3$ to make our experimental conditions comparable among the three datasets.

\pgfplotstableread{
name our    their
(1)   0.67    0.33
(2)   0.71    0.29
(3)   0.37    0.63
(4)   0.69    0.31
(5)   0.38    0.62
(6)   0.79    0.21
(7)   0.70    0.30
(8)   0.41    0.59
(9)   0.77    0.23
(10)  0.67    0.33
}\datatable
\begin{figure}
  \centering
  \scriptsize
  \setlength{\tabcolsep}{2pt}
  \begin{tabular}{*{5}{c}}
    \includegraphics[width=0.1\linewidth, trim=20 30 20 20, clip=true]{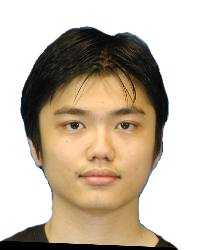} &
    \includegraphics[width=0.1\linewidth, trim=20 30 20 40, clip=true]{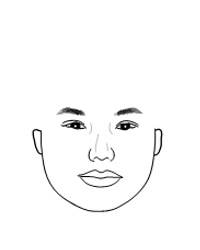} &
    \includegraphics[width=0.1\linewidth, trim=20 30 20 40, clip=true]{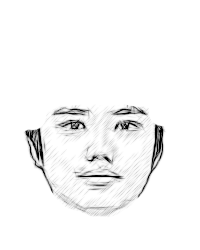} &
    \includegraphics[width=0.1\linewidth, trim=17 28 17 40, clip=true]{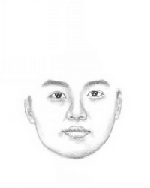} &
    \includegraphics[width=0.1\linewidth, trim=20 30 20 40, clip=true]{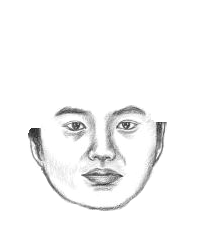} \\
    \includegraphics[width=0.1\linewidth, trim=20 20 10 20, clip=true]{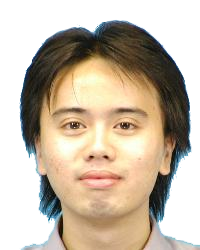} &
    \includegraphics[width=0.1\linewidth, trim=20 20 20 40, clip=true]{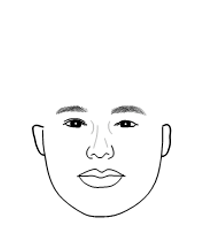} &
    \includegraphics[width=0.1\linewidth, trim=20 20 20 40, clip=true]{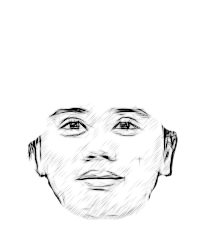} &
    \includegraphics[width=0.1\linewidth, trim=15 17 15 40, clip=true]{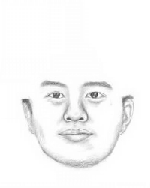} &
    \includegraphics[width=0.1\linewidth, trim=20 20 20 40, clip=true]{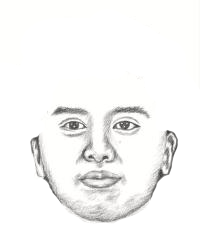} \\
    \includegraphics[width=0.1\linewidth, trim=18 20 22 40, clip=true]{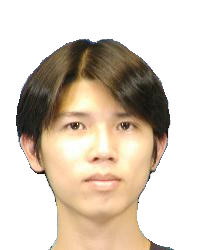} &
    \includegraphics[width=0.1\linewidth, trim=20 20 20 40, clip=true]{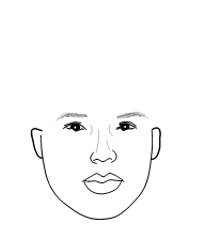} &
    \includegraphics[width=0.1\linewidth, trim=20 20 20 40, clip=true]{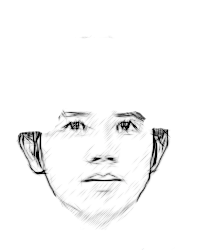} &
    \includegraphics[width=0.1\linewidth, trim=18 20 18 40, clip=true]{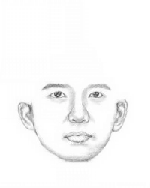} &
    \includegraphics[width=0.1\linewidth, trim=20 20 20 40, clip=true]{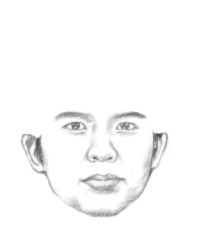} \\
    \includegraphics[width=0.1\linewidth, trim=20 20 20 20, clip=true]{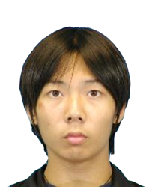} &
    \includegraphics[width=0.1\linewidth, trim=20 20 20 40, clip=true]{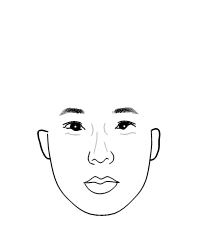} &
    \includegraphics[width=0.1\linewidth, trim=20 20 20 40, clip=true]{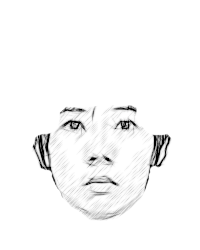} &
    \includegraphics[width=0.1\linewidth, trim=17 17 17 40, clip=true]{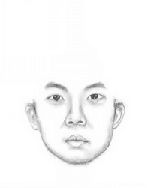} &
    \includegraphics[width=0.1\linewidth, trim=20 20 20 40, clip=true]{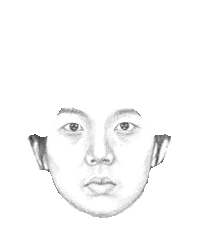} \\
    Photo & (a) & (b) & (c) & Artist I \\
  \end{tabular}
  \caption{ Comparing our approach with two commercial software packages. Column (a) and (b) are generated by the two commercial software packages~\cite{siyanhui} and \cite{akvis2012} respectively. Column (c) is generated by our approach with the training examples from $\Pi_1$. The sketch illustrations created by artist I are the groundtruths, which are not involved in the training process.}
  \label{fig:cmp_rslt}

\includegraphics[width=0.5\linewidth]{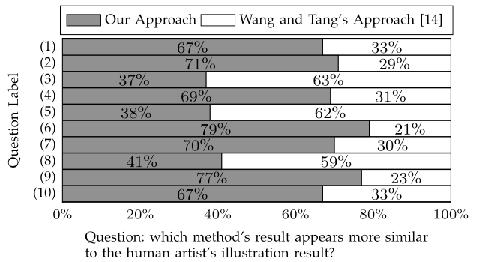}
\caption{The user study comparing the sketch generation quality by our approach and~\cite{Wang2009}. Data collected from 2624 completed questionnaires by answering 10 questions.}
\label{fig:sub_rslt}
\end{figure}

\begin{figure*}
    \center
    \scriptsize
    \setlength{\tabcolsep}{2pt}
    \renewcommand{\arraystretch}{0.5}
    \begin{tabular}{*{12}{c}}
        \includegraphics[width=0.072\linewidth, trim=20 20 20 20, clip=true]{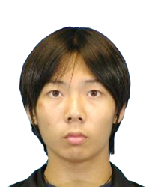} &
        \includegraphics[width=0.072\linewidth, trim=21 22 21 40, clip=true]{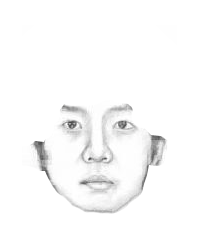} &
        \includegraphics[width=0.072\linewidth, trim=18 15 18 40, clip=true]{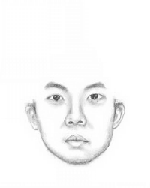} &
        \includegraphics[width=0.072\linewidth, trim=20 20 20 40, clip=true]{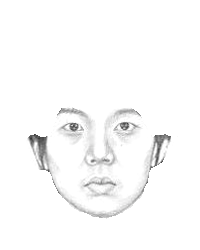} &
        \includegraphics[width=0.072\linewidth, trim=10 10 10 20, clip=true]{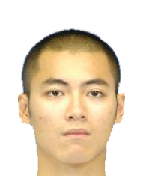} &
        \includegraphics[width=0.072\linewidth, trim=10 15 10 20, clip=true]{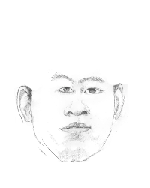} &
        \includegraphics[width=0.072\linewidth, trim=10 10 10 20, clip=true]{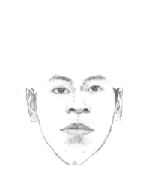} &
        \includegraphics[width=0.072\linewidth, trim=10 10 10 20, clip=true]{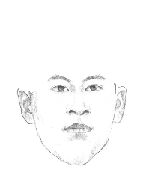} &
        \includegraphics[width=0.072\linewidth, trim=10 12 10 18, clip=true]{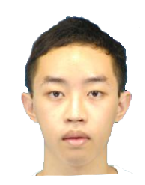} &
        \includegraphics[width=0.072\linewidth, trim=10 10 10 20, clip=true]{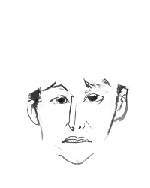} &
        \includegraphics[width=0.072\linewidth, trim=10 0 10 30, clip=true]{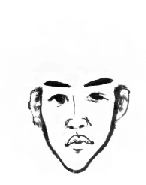} &
        \includegraphics[width=0.072\linewidth, trim=10 0 10 30, clip=true]{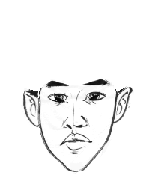} \\
        \includegraphics[width=0.072\linewidth, trim=20 20 20 30, clip=true]{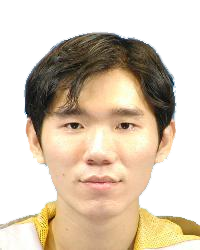} &
        \includegraphics[width=0.072\linewidth, trim=15 20 15 35, clip=true]{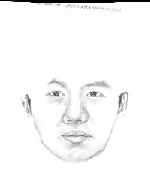} &
        \includegraphics[width=0.072\linewidth, trim=20 20 10 30, clip=true]{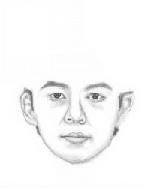} &
        \includegraphics[width=0.072\linewidth, trim=20 25 20 30, clip=true]{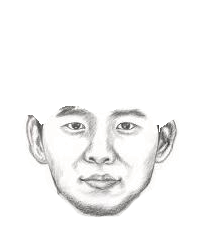} &
        \includegraphics[width=0.072\linewidth, trim=15 10 5 10, clip=true]{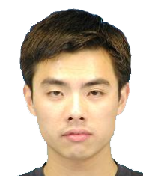} &
        \includegraphics[width=0.072\linewidth, trim=10 15 10 20, clip=true]{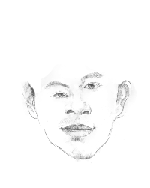} &
        \includegraphics[width=0.072\linewidth, trim=10 10 10 20, clip=true]{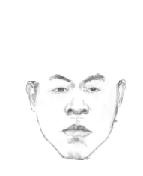} &
        \includegraphics[width=0.072\linewidth, trim=10 10 10 20, clip=true]{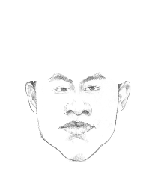} &
        \includegraphics[width=0.072\linewidth, trim=10 10 10 10, clip=true]{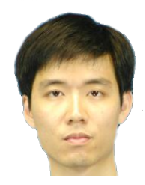} &
        \includegraphics[width=0.072\linewidth, trim=10 10 10 20, clip=true]{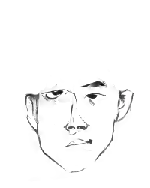} &
        \includegraphics[width=0.072\linewidth, trim=10 0 10 30, clip=true]{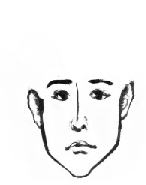} &
        \includegraphics[width=0.072\linewidth, trim=10 0 10 30, clip=true]{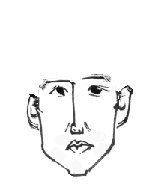} \\
        \includegraphics[width=0.072\linewidth, trim=20 20 20 20, clip=true]{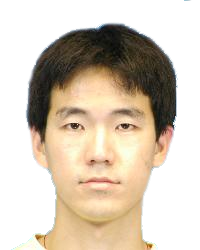} &
        \includegraphics[width=0.072\linewidth, trim=15 15 15 35, clip=true]{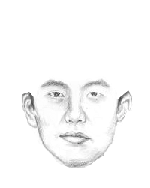} &
        \includegraphics[width=0.072\linewidth, trim=17 20 17 30, clip=true]{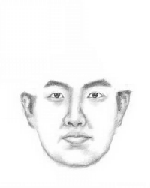} &
        \includegraphics[width=0.072\linewidth, trim=20 20 20 30, clip=true]{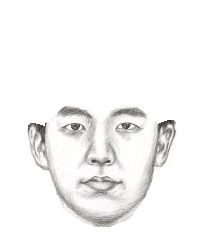} &
        \includegraphics[width=0.072\linewidth, trim=15 10 5 20, clip=true]{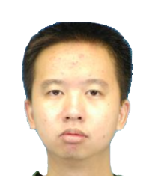} &
        \includegraphics[width=0.072\linewidth, trim=15 15 5 20, clip=true]{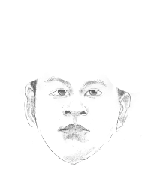} &
        \includegraphics[width=0.072\linewidth, trim=15 10 5 20, clip=true]{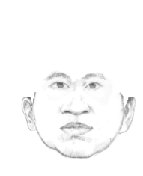} &
        \includegraphics[width=0.072\linewidth, trim=10 10 10 20, clip=true]{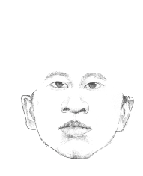} &
        \includegraphics[width=0.072\linewidth, trim=10 10 10 20, clip=true]{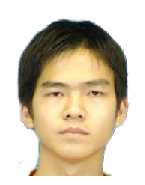} &
        \includegraphics[width=0.072\linewidth, trim=10 10 10 20, clip=true]{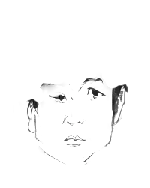} &
        \includegraphics[width=0.072\linewidth, trim=10 0 10 30, clip=true]{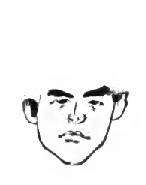} &
        \includegraphics[width=0.072\linewidth, trim=10 0 10 30, clip=true]{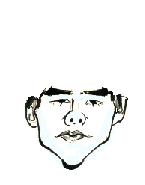} \\
        \includegraphics[width=0.072\linewidth, trim=15 20 25 10, clip=true]{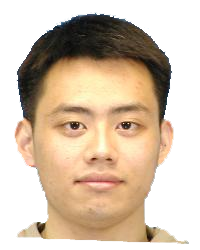} &
        \includegraphics[width=0.072\linewidth, trim=10 20 20 35, clip=true]{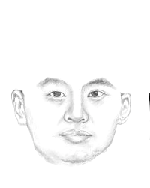} &
        \includegraphics[width=0.072\linewidth, trim=15 20 15 30, clip=true]{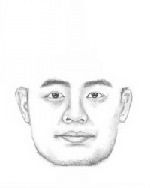} &
        \includegraphics[width=0.072\linewidth, trim=20 20 20 30, clip=true]{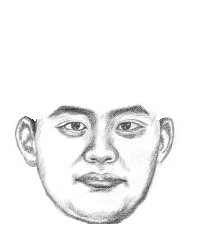} &
        \includegraphics[width=0.072\linewidth, trim=10 5 15 10, clip=true]{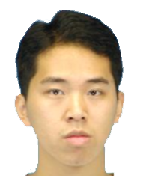} &
        \includegraphics[width=0.072\linewidth, trim=10 15 15 20, clip=true]{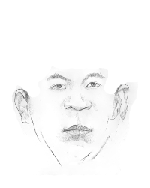} &
        \includegraphics[width=0.072\linewidth, trim=10 5 15 20, clip=true]{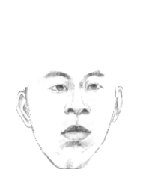} &
        \includegraphics[width=0.072\linewidth, trim=10 5 15 20, clip=true]{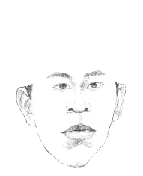} &
        \includegraphics[width=0.072\linewidth, trim=10 0 10 15, clip=true]{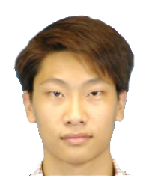} &
        \includegraphics[width=0.072\linewidth, trim=10 0 10 30, clip=true]{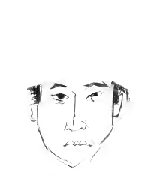} &
        \includegraphics[width=0.072\linewidth, trim=10 0 10 30, clip=true]{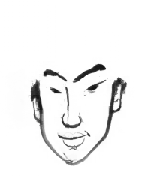} &
        \includegraphics[width=0.072\linewidth, trim=10 0 10 30, clip=true]{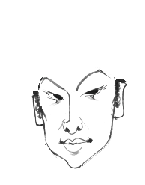} \\
        \includegraphics[width=0.072\linewidth, trim=20 20 20 30, clip=true]{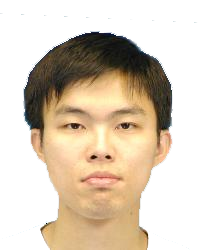} &
        \includegraphics[width=0.072\linewidth, trim=15 15 15 35, clip=true]{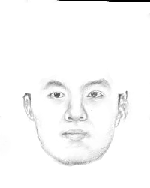} &
        \includegraphics[width=0.072\linewidth, trim=17 20 17 30, clip=true]{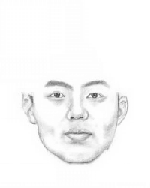} &
        \includegraphics[width=0.072\linewidth, trim=20 20 20 30, clip=true]{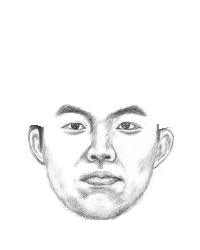} &
        \includegraphics[width=0.072\linewidth, trim=10 10 10 20, clip=true]{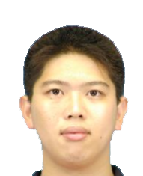} &
        \includegraphics[width=0.072\linewidth, trim=10 20 10 20, clip=true]{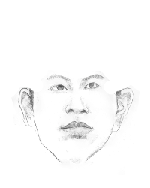} &
        \includegraphics[width=0.072\linewidth, trim=10 10 10 20, clip=true]{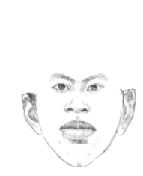} &
        \includegraphics[width=0.072\linewidth, trim=10 10 10 20, clip=true]{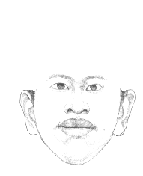} &
        \includegraphics[width=0.072\linewidth, trim=10 0 10 20, clip=true]{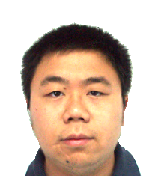} &
        \includegraphics[width=0.072\linewidth, trim=10 0 10 30, clip=true]{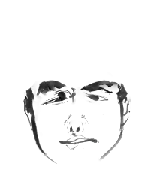} &
        \includegraphics[width=0.072\linewidth, trim=10 0 10 30, clip=true]{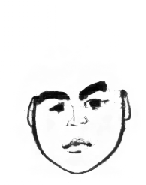} &
        \includegraphics[width=0.072\linewidth, trim=10 0 10 30, clip=true]{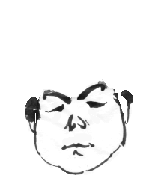} \\
        \includegraphics[width=0.072\linewidth, trim=15 20 15 20, clip=true]{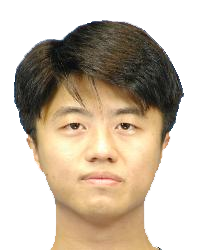} &
        \includegraphics[width=0.072\linewidth, trim=15 20 15 30, clip=true]{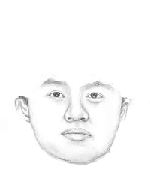} &
        \includegraphics[width=0.072\linewidth, trim=15 20 15 30, clip=true]{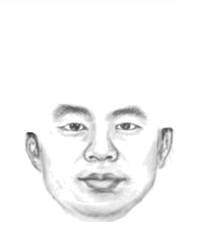} &
        \includegraphics[width=0.072\linewidth, trim=15 20 15 30, clip=true]{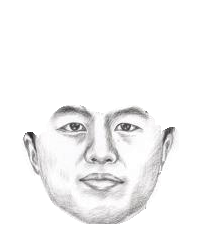} &
        \includegraphics[width=0.072\linewidth, trim=10 10 10 20, clip=true]{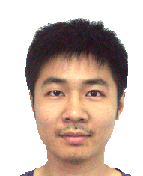} &
        \includegraphics[width=0.072\linewidth, trim=10 20 10 20, clip=true]{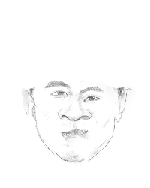} &
        \includegraphics[width=0.072\linewidth, trim=15 10 5 20, clip=true]{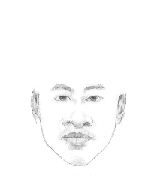} &
        \includegraphics[width=0.072\linewidth, trim=10 10 10 20, clip=true]{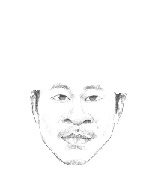} &
        \includegraphics[width=0.072\linewidth, trim=10 10 5 25, clip=true]{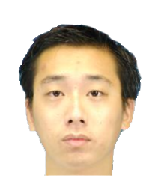} &
        \includegraphics[width=0.072\linewidth, trim=10 10 5 25, clip=true]{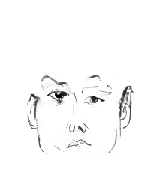} &
        \includegraphics[width=0.072\linewidth, trim=10 0 10 25, clip=true]{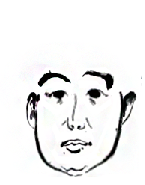} &
        \includegraphics[width=0.072\linewidth, trim=10 0 10 30, clip=true]{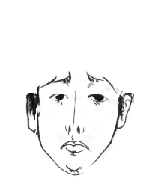} \\
        Photo & (a) & (b) & Artist I & Photo & (a) & ~(b) & Artist II & Photo & (a) & ~(b) & Artist III\\
    \end{tabular}
    \caption{Generating sketch illustrations of facial portraits from input human facial photos in personal illustration styles of three artists. Column (a) are generated by the peer algorithm proposed in~\cite{Wang2009} while column (b) are  produced by our algorithm.}
    \label{fig:gen_rslt}
\end{figure*}
To compare the aesthetic stylization between our approach with that of a facial sketch synthesis approach in~\cite{Wang2009} and two commercial packages~\cite{siyanhui,akvis2012}, we perform the leave-one-out test on $\Pi_1$. It is noted that since the sketch illustrations in $\Pi_1$ do not present the exaggerative style, the procedure of removing exaggeration from example sketches (Sec.~\ref{sec:IUS}) is not performed. That is, for $\Pi_1$, $S_i$ and $S_i'$ are the same. Fig.~\ref{fig:cmp_rslt} shows the results by our method and the two peer commercial software packages, which clearly demonstrate that the sketch illustrations generated by us appear visually much closer to the target artist's hand-drawn ones. Fig.~\ref{fig:gen_rslt} gives more sketching instances by our approach and the state-of-the-art peer algorithmic approach \cite{Wang2009}, which also intuitively leads to the same qualitative comparison conclusion. In this experiment, as we perform the leave-one-out test on $\Pi_1$, the number of training samples used for sketch generation is 48, which is significantly less than the number of sample sketches, at least 88, in \cite{Wang2009}. In all experiments in this paper, less than 50 training samples are used to capture an artist's personal facial sketch illustration style.

To make more assessment on the stylization of the sketches produced by our approach and the peer algorithm, we further conducted a user survey on the sketches generated in the above experiment. We selected 10 artist-drawn sketches from $\Pi_1$ and their corresponding sketches generated by our approach and \cite{Wang2009} respectively. We then created an online questionnaire consisting of 10 questions. Each question presents three images side by side. The hand-drawn sketch illustration is always shown in the middle. The corresponding sketch illustrations generated by the two algorithms respectively are randomly placed on the left and right. Each question asks a human viewer to select among the two images displayed on the left and right positions of the image triplet. The subjects are asked to answer which one appears visually more similar to the middle image.
We released the online questionnaire on a twitter-like social media community dedicated to comic fans~\cite{sinaweibo}, who are generally very familiar with artistic facial drawing styles. 2624 completed questionnaires were collected
during 3 days. None of the participants was compensated monetarily, who took part in the online survey due to their curiosity and interest in facial sketch illustrations. Through a dynamic webpage tracking feature, it is shown each questionnaire takes around 2.5 minutes on average to be answered. Among the ten pairs of sketches in our online survey, seven of them get more votes for our sketches by the subjects, and the remaining three get more votes for the sketches from the peer algorithm. Overall, $62\%$ of all the 26240 answers to the ten questions favor resulting sketches by our method (one-sample t-test, $p\text{-}value < 0.001$; two-sample t-test, $p\text{-}value < 0.001$). Fig.~\ref{fig:sub_rslt} presents more details about our assessment. The voting clearly shows the superiority of our method in terms of the visual appearance of sketching.

We also perform leave-one-out tests on the data sets $\Pi_2$ and $\Pi_3$. Fig.~\ref{fig:gen_rslt} gives more synthesized sketches, which demonstrate that our approach is indeed capable of imitating multiple personal facial illustration styles, with or without exaggerations.
The first artist sketches human face portraits more or less following the visual characteristics in the original facial photo. The other two artists present significant exaggeration in their facial illustrations.

In Fig.~\ref{fig:gen_rslt}, it is obvious that the peer algorithm~\cite{Wang2009} fails to capture the sketch illustration styles of artists II and III as the generated sketch illustrations have significant differences to the ones created by the target artists. This is partly because their algorithm is designed upon the assumption that each facial region in a photo occupies the same image area as its counterpart region in the corresponding sketch illustration does, which is not always true in reality, particularly for sketch illustrations with significant exaggeration. 

\begin{figure}
    \center
    \scriptsize
    \includegraphics[width=0.7\linewidth]{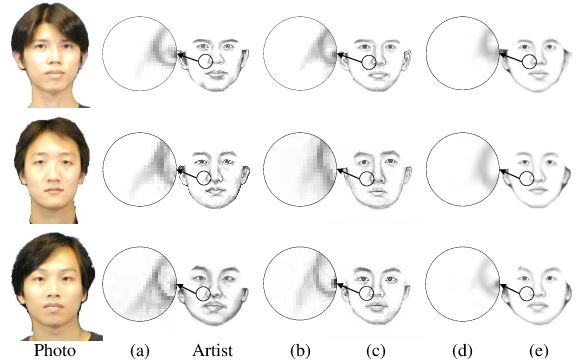}
    \caption{Comparing our approach with the method proposed by~\cite{wang2013transductive}. The faces in column (c) and (e) are generated by our algorithm and the peer algorithm proposed in~\cite{wang2013transductive} respectively.}
    \label{fig:cmp_trans_rslt}
    \setlength{\tabcolsep}{0pt}
    \begin{tabular}{*{5}{c}}
        \includegraphics[width=0.1\linewidth, trim=10 10 10 20, clip=true]{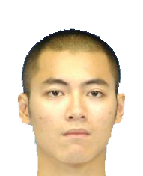} &
        \includegraphics[width=0.1\linewidth, trim=10 20 10 50, clip=true]{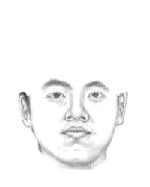} &
        \includegraphics[width=0.1\linewidth, trim=10 25 10 45, clip=true]{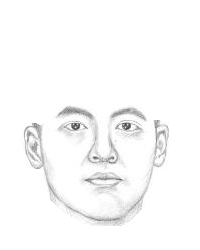} &
        \includegraphics[width=0.1\linewidth, trim=10 10 10 60, clip=true]{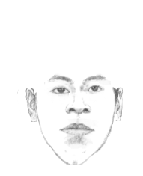} &
        \includegraphics[width=0.1\linewidth, trim=10 10 10 60, clip=true]{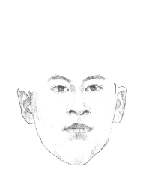} \\
        \includegraphics[width=0.1\linewidth, trim=13 10 7 7, clip=true]{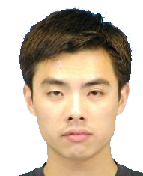} &
        \includegraphics[width=0.1\linewidth, trim=15 20 15 60, clip=true]{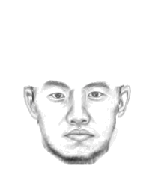} &
        \includegraphics[width=0.1\linewidth, trim=20 30 20 60, clip=true]{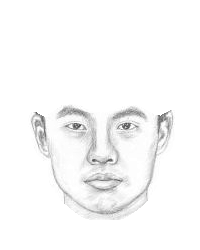} &
        \includegraphics[width=0.1\linewidth, trim=10 10 10 60, clip=true]{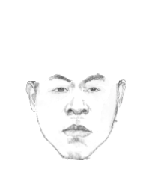} &
        \includegraphics[width=0.1\linewidth, trim=10 10 10 60, clip=true]{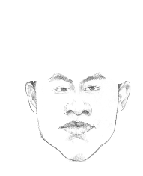} \\
        \includegraphics[width=0.1\linewidth, trim=15 10 5 20, clip=true]{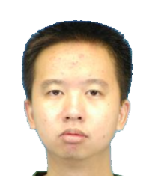} &
        \includegraphics[width=0.1\linewidth, trim=13 20 13 60, clip=true]{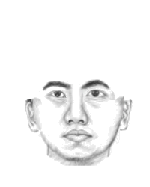} &
        \includegraphics[width=0.1\linewidth, trim=15 20 15 60, clip=true]{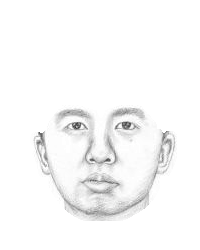} &
        \includegraphics[width=0.1\linewidth, trim=15 10 5 60, clip=true]{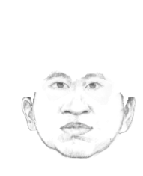} &
        \includegraphics[width=0.1\linewidth, trim=10 10 10 60, clip=true]{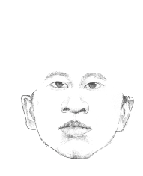} \\
        \includegraphics[width=0.1\linewidth, trim=12 5 8 25, clip=true]{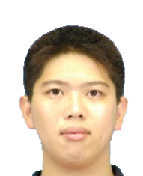} &
        \includegraphics[width=0.1\linewidth, trim=10 10 10 60, clip=true]{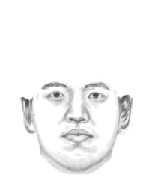} &
        \includegraphics[width=0.1\linewidth, trim=10 10 10 60, clip=true]{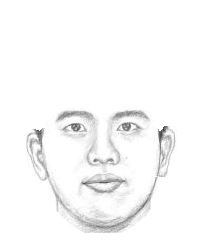} &
        \includegraphics[width=0.1\linewidth, trim=15 10 5 60, clip=true]{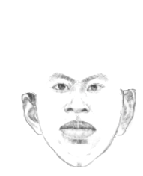} &
        \includegraphics[width=0.1\linewidth, trim=10 10 10 60, clip=true]{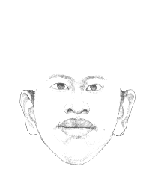} \\
        Photo & (a) & Artist 1 & (b) & Artist 2 \\
    \end{tabular}
    \caption{Human facial sketch portraits generated in two illustration styles for the same set of input facial photos. Images in column (a) and (b) are generated by our approach. The training data for generating the images in columns (a) and (b) come from $\Pi_1$ and $\Pi_2$ respectively.}
    \label{fig:gen_rslt_from_same_photos}
\end{figure}

We also compare our approach with the transductive learning algorithm proposed by Wang et al.~\cite{wang2013transductive}.
Fig.~\ref{fig:cmp_trans_rslt} shows our resultant sketches in the leave-one-out test on $\Pi_1$ and sketches from Wang's website of results on CUHK face sketch database~\cite{CUHKFace2013}.
The nasolabial folds~\cite{burgess2005cosmetic} is one of the essential elements of facial stylization and often appears in face portraits drawn by human artist. They are well preserved and can be significantly observed in our resultant sketches. Unfortunately, nasolabial folds are little presented in sketches generated by the transductive learning algorithm. Hence, our resultant sketches more closely resembles the groundtruth sketches by the target human artists.

Fig.~\ref{fig:gen_rslt_from_same_photos} shows more resulting sketches in the leave-one-out tests on $\Pi_2$ and $\Pi_3$. It demonstrates that our method can successfully learn personal sketching styles of multiple artists and accordingly generate the individual stylized sketches for the same input photos.

Regarding the time performance, it takes about 1 minute to generate a facial sketch from an input photo of the resolution of 200 by 250 pixels using the unoptimized, single computing core prototype implementation of our algorithm executed on a PC equipped with Intel i5-3450 3.1GHz CPU and 3.2G memory, while the peer algorithm~\cite{Wang2009} takes 3 minutes on the same hardware platform.

Originally our training takes about 3 days to finish. However, we observe that it is straightforward to implement our training algorithm concurrently since FE and PE models for each facial region can be trained independently.
We evaluated the parallel computing implementation on a PC with Intel i7 CPU (8 cores) and 6G memory.
It took about 8 hours to complete all training tasks. We are thus optimistic about the computational efficiency of our algorithm in practice with a parallel implementation.

\section{Conclusion and Discussions}

We propose a new CBR-based sketch synthesis algorithm that can produce visually superior results than existing synthesis methods. To the best of our knowledge, it is the first CBR framework that is explicitly introduced to personally stylize human facial portrait.

For each human artist to be mimicked, a series of cases are firstly built-up
from her/his exemplars of source facial photo and hand-drawn sketch, and then its stylization for facial photo is transformed as a style-transferring process of iterative refinement by looking for and applying a series of best-sit cases in a sense of style optimization.
The presented experimental results demonstrate that in comparison with a state-of-the-art method and a couple commercially available software packages, the new approach is capable of generating visually more appealing portraits from the point of view of personal style, which also more closely resemble the groundtruth sketches by the target human artists.

Our method has a major limitation. It has not supported the synthesize sketch illustrations for human hair yet. This limitation comes from the difficulty in establishing correspondences between strokes or texture areas that depict hairs and the counterpart hair regions on human facial photos. A related future research direction of this problem is Cross-Modal Face Matching~\cite{ouyang2014face}. Besides, if we focus on the hair of a specific group of people, it is possible to represent the correspondences through manually indicated key points, like that by Chen et al.~\cite{Chen2004}.

\end{document}